# Algorithms for synthesis of three-dimensional warehouse systems configurations optimal in terms of minimum cost and maximum speed

A.V. Razumovsky, M.V. Saramud, S.B. Tkachev, N.V. Shtabel

Siberian State University of Science and Technology named after Academician M.F. Reshetnev, Russian Federation, 660037, Krasnoyarsk, Krasnoyarskiy rabochiy avenue, 31

***Abstract:*** *Algorithms for creating optimal configurations of warehouse three-dimensional systems consisting of unified transport and warehouse modules are presented in the article. The constructions of these modules, and methods for constructing a warehouse array map are described. Each module has a cubic shape and is connected to each other with any of the 6 faces. In order to save money, some of the modules do not have drives to move the load in the vertical direction. When building a warehouse, a problem to determine the optimal warehouse configuration in terms of cost-performance ratio arises.*

It is proposed to consider warehouse systems as a control object built on the basis of unified transport and storage cells, each of which is a storage place and is equipped with a mechanism for moving cargo in one of six directions. Figure 1 shows a schematic diagram of a transport and storage cell.

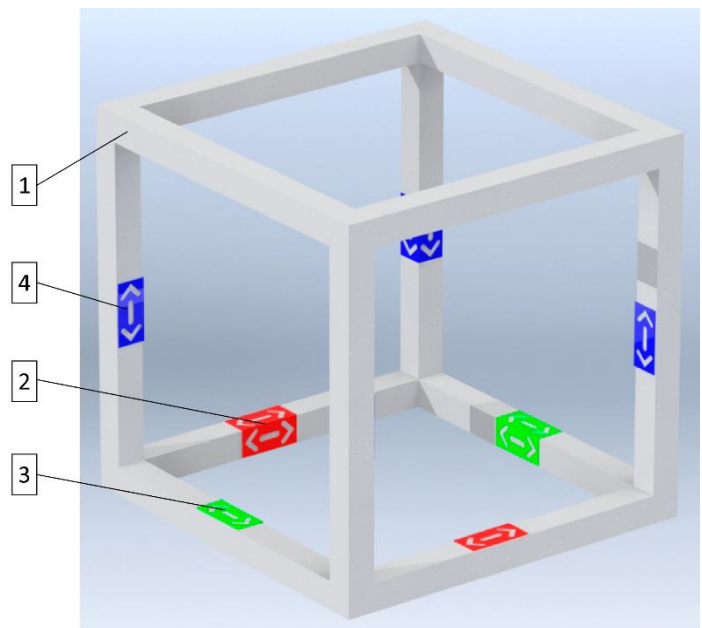

Figure 1 - Scheme of a unified transport and storage cell

The cell consists of a fixed frame 1, in which there are devices for fixing the load (not shown in the diagram), as well as a drive for moving loads 2, 3 and 4: along the X axis (2 pcs.), along the Y axis (2 pcs.), along Z axis (4 pcs.). Drives along each of the coordinate axes operate synchronously and can move the load to one of the six neighboring cells (Figure 2). A more economical cell configuration does not have drives along the Z axis.

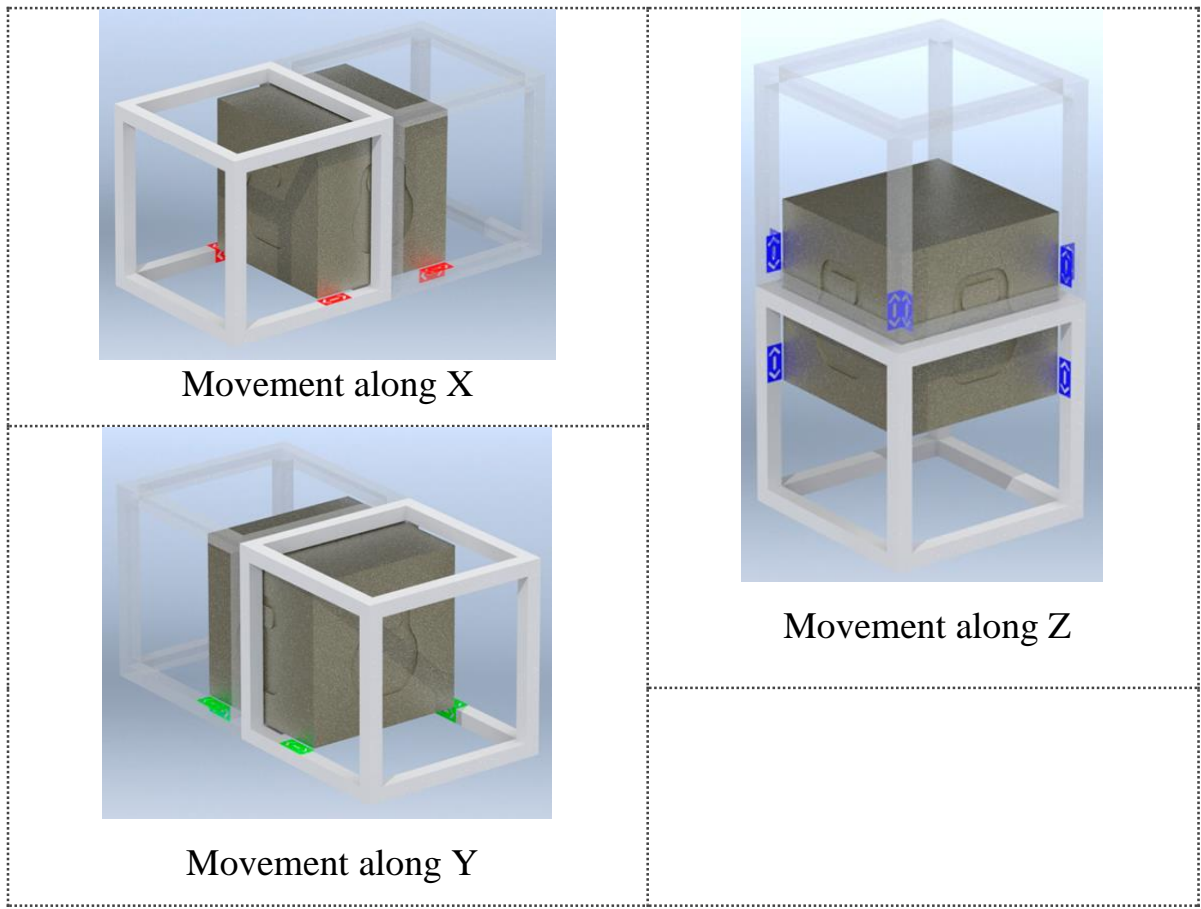

Figure 2 - Options for moving goods along the coordinate axes

**Communication between cells and addressing**

Each warehouse cell is an autonomous device with its own managing controller. When assembling such cells into a three-dimensional array, it is necessary to determine the location of each cell in the array and the configuration of the entire array. As a result, each cell receives a unique address in the array and a map of the cell array is built.

It is inappropriate to use bus, star, or delta topologies for communication between cells, since they do not make it possible to determine the physical location of the cells relative to each other. Therefore, a mesh topology (mesh network) was chosen to organize data exchange [1], similar to the one used in wireless mesh networks. The difference from wireless networks is a fixed number of connections with neighboring cells (6 channels, one for each face of a cubic cell), so this topology can be called a hex mesh network.

Usage of this approach allows one to determine from which side the information physically comes, and accordingly build a network map. As a communication interface, the UARTS interface available in almost any microcontroller is used. [2] or its interference-free alternative (RS232, RS422, …) [3]. Since the communication channel is organized only between neighboring cells, differential data transmission is not required, it is advisable to use UART or RS232 interfaces to minimize the number of contacts in the connector. Further we will assume the use of the UART interface. The UART interface is asynchronous, which allows one to simultaneously receive and transmit data on each of the interfaces.

Thus, cells are able to transmit information to each other, using two message passing mechanisms: direct distribution (flooding) and addressing (routing).

**Cells addressing**

Being powered up for the first time, the cells need to map the warehouse and determine their location in it, as well as a unique address. For this purpose, special messages are used, transmitted by the flooding method. When the power is turned on, each cell issues to its interfaces UART1...UART6 (corresponding to the edges 1...6 of the cell) an addressing packet containing only the cell address, which is formed from a unique identifier programmed in the microcontroller at the producing factory (for STM32 microcontrollers 96 bits) [4].

The addressing packets arriving at the input of the UART1 ... UART6 interfaces must be supplemented with their address and send them further to all interfaces, except for the one from which the packet came [5].

Thus, following through the array of cells, the addressing packet is supplemented with the addresses of all the cells through which it passes, and a unique route is formed. In this case, own packets can be identified by the first address, which must match the own one.

In case the received packet met its own address, and this is its own packet, one must save this packet as a route. In the future, the packet data can be used to address any of the cells that occur along the route.

If the packet is not its own and its address occurs in the packet for the second time, it is necessary to discard such a packet and not transmit it further, since this route is a crisscross.

Thus, the received packets with completed routes make it possible to evaluate both possible routes and existing links between neighboring cells. This allows one to build a map / tree / graph of cells, for processing by routing algorithms.

When addressing packets are sent by cell 1, the following routes will be detected:

1 - 2 - 3 - 6 - 9 - 8 - 7 - 4 - 1 and its pair 1 - 4 - 7 - 8 - 9 - 6 - 3 - 2 - 1

1 - 2 - 3 - 6 - 5 - 4 - 1 and doubles 1 - 4 - 5 - 6 - 3 - 2 - 1

1 - 2 - 5 - 8 - 7 - 4 - 1 and doubles 1 - 4 - 7 - 8 - 5 - 2 - 1

1 - 2 - 5 - 4 - 1 and doubles 1 - 4 - 5 - 2 - 1

1 - 2 - 3 - 6 - 9 - 8 - 5 - 4 -1 and doubles 1 - 4 - 5 - 8 - 9 - 6 - 3 - 2 - 1

1 - 2 - 3 - 6 - 5 - 8 - 7 - 4 - 1 and doubles 1 - 4 - 7 - 8 - 5 - 6 - 3 - 2 – 1

while intersecting routes

1 - 2 - 5 - 6 - 3 - 2 - 1 and 1 - 4 - 5 - 8 -7 - 4 – 1

will not be taken into account, since the passage through one cell is carried out twice.

The map can be calculated and stored both in each cell controller and in one main controller, for example, at the point of receiving / issuing goods. Each of the methods has its own advantages and disadvantages.

If it is necessary to redirect the cell (for example, after a reboot), the cell initiates the routing process on its own, starting the distribution of address packets over all interfaces.

**Design and execution of a compound cell of a flexible warehouse**

The modular cell of a flexible warehouse structurally consists of the main elements: a frame 1, a container 2, a drive of horizontal 3 and vertical 4 movements, allowing the container 2 to be moved between adjacent cells.

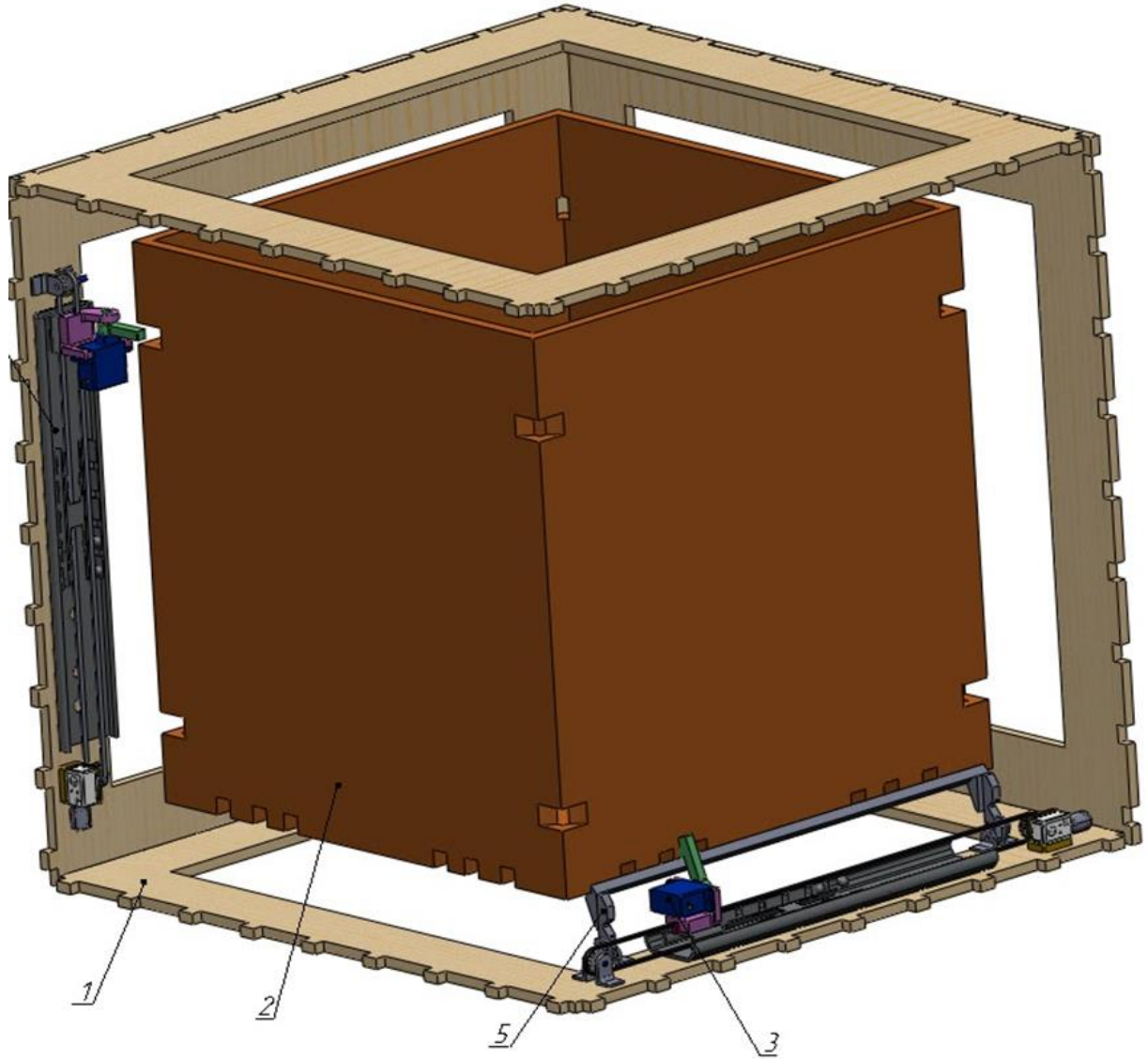

Figure 3 - Structural implementation of the drive for vertical and horizontal movement of the container

Drives 3 and 4 displacements are identical and differ only in the length of execution. The support frame 5 serves to hold the container 2 inside the frame 1 and at the same time to center the container inside the frame and also to set the direction of movement.

The unified displacement drive in figure 3 consists of a gear motor 1, which, through a toothed belt drive, moves the carriage 3 along the guide 2 with the servo drive 4 installed on it, which rotates the pusher 5. The pusher 5 serves to transfer force from the drive to the container 2 (Figure 5.4) while simultaneously raising and lowering the support frame 5 (Figure 5.5).

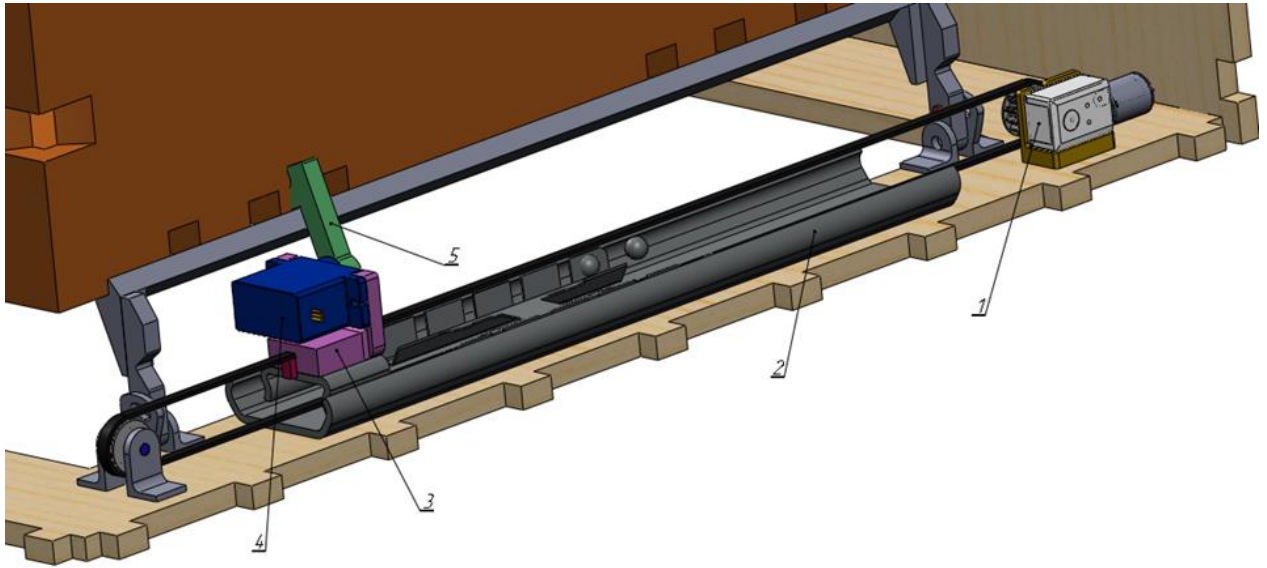

Figure 4 - Unified displacement drive

**Warehouse array indicators**

To speed up calculations, a three-dimensional array warehouse with the number of modules 4x4x3, with a fixed area for loading and issuing goods (in the lower corner of the array) is further considered.

At the stage of forming the optimal configuration, the sum of all movements necessary for the sequential loading of the entire warehouse acts as an indicator of the speed of the warehouse. It means, the first load enters the loading area of the empty warehouse, which moves to the diagonally opposite corner. Then the second load that has entered the loading area is moved to the far corner next to the first load, and so on.

As a cost, it was assumed that one module capable of moving cargo along three coordinate axes (hereinafter referred to as the "three-axis module") consists of: frame and related mechanical parts - 0.2 c.u.
    drives along the X axis - 0.2 c.u.
    drives along the Y axis - 0.2 c.u.
    drives along the Z axis - 0.4 c.u.

Thus, the total cost of such a module is 1 c.u., respectively, a module without the possibility of vertical movement (two-axis module) will cost 0.6 c.u.

**Optimization objective function**

When optimizing, one should strive to minimize the sum of all movements necessary to load the warehouse and to the minimum cost of the warehouse array.

Obviously the fastest, but also the most expensive, would be a warehouse array consisting of only three-axis modules. Another extreme case is a cheap but slow warehouse that has only three 3-axis cells (one at each level) and the rest of the modules are 2-axis.

In a real situation, the two considered extreme cases correspond to warehouses, for example, in a grocery store, and in the second, a warehouse for storing personal belongings, where requests can be significantly smaller.

Thus, in order to link these two criteria into a single objective function, this kind of formula was used:

$$F_{target}(F_{speed}, F_{cost}, \alpha) = \alpha \cdot \frac{F_{speed}}{F_{speed.n}} + (1-\alpha) \cdot \frac{F_{cost}}{F_{cost.n}} \to min$$

where, α is the coefficient of significance of the speed criterion in relation to the cost criterion α∈[0;1];

$F_{speed}$ and $F_{speed.n}$ – storage array speed criterion and its normalizing value (the sum of all movements required to sequentially load the entire warehouse);

$F_{cost}$ and $F_{cost.n}$ – warehouse array cost criterion and its normalizing value (the sum of the costs of all cells in the warehouse array).

The sum of movements during loading corresponding to the warehouse with all three-axis cells and the cost of all its cells were taken as normalizing values:
$$F_{speed.n} = 7808 \text{ и } F_{cost.n} = 48$$

**Analysis of the warehouse array structure**

Having carried out the optimization with the coefficient of significance of the warehouse speed α=1, the structure shown in Figure 5 ($F_{target} = 1$) will be obtained. In the warehouse array graphics, two-axis cells (which can only move the load along the *xy* axes) are shown in green, and three-axis cells (which can move the load along the *xyz* axes) are shown in blue.

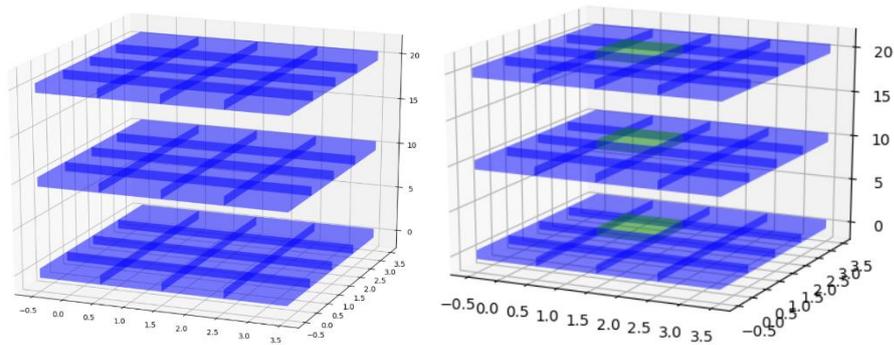

Figure 5 - Two best warehouse configurations with α=1

The second option in terms of characteristics is the presence of one column of *xy* cells anywhere in the warehouse ($F_{speed} = 7820$, $F_{cost} = 46.8$, $F_{target} = 1,002$).

($F_{speed} = 8144$, $F_{cost} = 34.8$, $F_{target} = 0.884$) и ($F_{speed} = 8360$, $F_{cost} = 33.6$, $F_{target} = 0.885$)

With equal significance of both criteria (i.e. α=0.5), the two best configurations presented in Figure 6 were obtained

($F_{speed} = 8144$, $F_{cost} = 34.8$, $F_{target} = 0.884$) и ($F_{speed} = 8360$, $F_{cost} = 33.6$, $F_{target} = 0.885$)

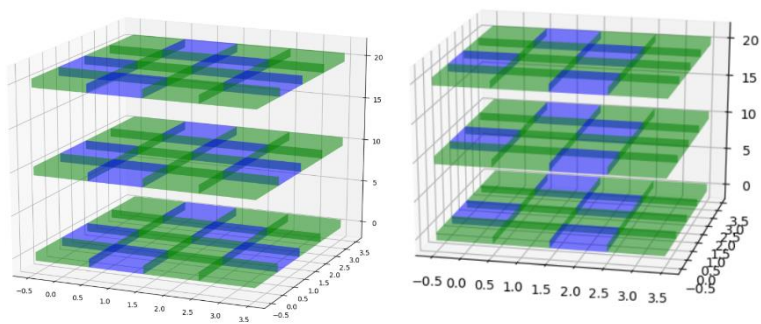

Figure 6 - Two best configurations at α=0.5

With the essential significance of the warehouse cost criterion (α=0.1), two warehouse configurations were obtained, which are shown in Figure 7. In each of the two configurations, the number of three-axis cells is 6 pieces, but their location is different ($F_{speed} = 8960$, $F_{cost} = 31.2$, $F_{target} = 0.7$).

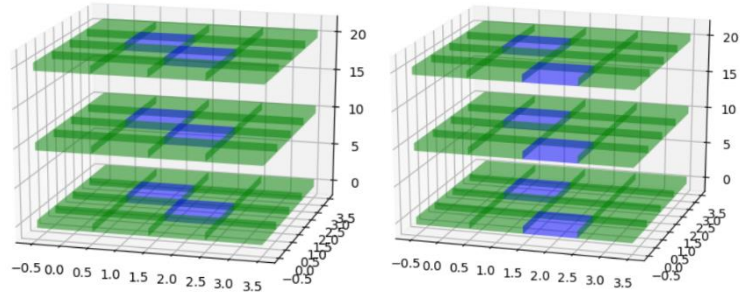

Figure 7 - Two best configurations at α=0.1

Table 1 shows all 5 warehouse configurations with calculated objective function values depending on the significance coefficient factor α.

Table 1 - Influence of significance coefficient factor α on the objective function

| Number of triaxial cells | F speed | F cost | F target |
|---|---|---|---|
| α=1 | | | |
| 48 | 7808 | 48 | 1 |
| 45 | 7820 | 46,8 | 1,002 |
| 15 | 8144 | 34,8 | 1,043 |
| 12 | 8360 | 33,6 | 1,071 |
| 6 | 8960 | 31,2 | 1,148 |
| α=0.5 | | | |
| 48 | 7808 | 48 | 1 |
| 45 | 7820 | 46,8 | 0,988 |
| 15 | 8144 | 34,8 | 0,884 |
| 12 | 8360 | 33,6 | 0,885 |
| 6 | 8960 | 31,2 | 0,899 |
| α=0.1 | | | |
| 48 | 7808 | 48 | 1 |
| 45 | 7820 | 46,8 | 0,978 |
| 15 | 8144 | 34,8 | 0,757 |
| 12 | 8360 | 33,6 | 0,737 |
| 6 | 8960 | 31,2 | 0,7 |

## Conclusion

A fundamentally new scheme for organizing storage systems based on unified transport and storage modules that have the ability to transfer cargo along three guides has been demonstrated. For the proposed scheme, the concept of organizing communication between the cells was developed and the design of the transport and storage module containing 8 unified drives was made. The developed algorithm for constructing work trajectories using a system of weights has demonstrated high flexibility and extensibility. The proposed complex objective function demonstrates its effectiveness in the formation of optimal warehouse structures.

The work was supported financially by the Ministry of Science and Higher Education of the Russian Federation (Agreement No. FEFE-2020-0017).